\documentclass[letterpaper]{article} 
\usepackage[]{aaai23}  
\usepackage{times}  
\usepackage{helvet}  
\usepackage{courier}  
\usepackage[hyphens]{url}  
\usepackage{graphicx} 
\usepackage{natbib}  
\usepackage{caption} 
\usepackage{algorithm}
\usepackage{algorithmic}
\usepackage{amsmath,amssymb}
\usepackage{booktabs}
\usepackage{multirow}
\usepackage{mathtools}
\usepackage[switch]{lineno}

\newcommand{\argmin}{\mathop{\arg\min}}

\newcommand{\sm}{\pi}

\usepackage{newfloat}
\usepackage{listings}
\DeclareCaptionStyle{ruled}{labelfont=normalfont,labelsep=colon,strut=off} 
\floatstyle{ruled}
\newfloat{listing}{tb}{lst}{}
\floatname{listing}{Listing}
\title{Detachedly Learn a Classifier for Class-Incremental Learning}
\author{
    Ziheng Li,
    Shibo Jie,
    Zhi-Hong Deng
}
\affiliations{
    School of Intelligence Science and Technology, Peking University\\
    \{liziheng, parsley, zhdeng\}@pku.edu.cn
}

\nocopyright

\begin{document}


\maketitle

\begin{abstract}
In continual learning, training a discriminative model need to continually learn a feature extractor and a classifier on a sequence of tasks. This paper empirically proves a pretrained feature extractor can significantly promote continual learning by means of a optimized experience replay (ER) strategy. We present an probabilistic analysis that the failure of vanilla ER comes from unnecessary re-learning of previous tasks and incompetence to distinguish current task from the previous ones, which is the cause of knowledge degradation and prediction bias. To overcome these weaknesses, we propose a novel replay strategy called task-aware experience replay. It rebalances the replay loss and detaches classifier weight for the old tasks from the update process, by which the previous knowledge is kept intact and the overfitting on episodic memory is alleviated. Experimental results show our method outperforms current state-of-the-art methods.
\end{abstract}

\section{Introduction}

Training of contemporary artificial neural networks heavily relies on the exposition to independent and identically distributed data sampling. When continually learning a sequence of tasks, neural network only works well on the most recent task and the performance on previous tasks drops drastically. This phenomenon is referred to as \textit{catastrophic forgetting}~\cite{1989cataf,1995cataf}. It restricts the application of neural networks in read-world scenarios. Continual learning (CL, a.k.a. incremental learning) aims to train a single model on non-stationary data distribution avoiding catastrophic forgetting with limited memory overhead~\cite{1989cataf,ewc}.

On the other hand, pretrained models have achieved great success in NLP and CV~\cite{bert,pt_nlp,vit}. Most state-of-the-art approaches for NLP downstream tasks rely on pretraining on large-scale dataset. For CV, this pretraining \& finetuning paradigm is also applied in continual learning to replace random initialization~\cite{agscl,cpr}. Some works keep the pretrained model unchanged to extract stable feature, on which they build a simple classifier~\cite{cilgc}, e.g. one-layer linear classifier. We think it is necessary to push the research of continual learning towards pretrained-model-based paradigm to achieve satisfactory performance in real-world applications. However, existing works still do not fully exploit the power of pretrained model. This paper aims to explore a effective way to continually learn a classifier upon a frozen feature extractor. Compared with finetuning, this linear probing paradigm also enjoys the pretrained knowledge but is much more efficient. 



Although pretrained feature extractor provides rich and stable feature, training the classifier still faces two challenges. First, the ability to classify among classes of past tasks degrades which we call knowledge degradation; Second, the classifier is unable to learn to distinguish between current task and past ones which causes the notorious prediction bias. Prediction bias happens because discriminative classifier needs both positive and negative samples to derive the decision boundary. Without previous data, the classifier cannot distinguish samples from different tasks. To this end, many class-incremental learning approaches maintain an episodic memory that stores a few samples from previous tasks to perform experience replay (ER) in following training~\cite{clear,eeil,ucir,gss,derpp}. 

In this paper, to explore a proper way that ER works with pretrained model, we first present an analysis from a probabilistic view that vanilla experience replay (ER) fails because of unnecessary re-learning of previous tasks and incompetence to distinguish the current task from the previous ones due to overfitting and class imbalance, which causes knowledge degradation and prediction bias respectively. To overcome the weaknesses of vanilla ER, we propose task-aware experience repaly (TaER). TaER differs from ER in three aspects. Firstly, TaER focuses on training classifier built on a frozen pretrained model. Secondly, TaER adds a dynamic weight factor to the replay loss term to mimic a class-balanced training. Thirdly, TaER only updates the classifier weight for the current task so that the old classifier is detached from the gradient flow (still involve in softmax), which completely eliminates the forgetting and significantly alleviates overfitting. In summary, our contribution are three-fold:
\begin{enumerate}
    \item We empirically prove a pretrained feature extractor can significantly promote continual learning.
    \item We propose a novel ER strategy which better exploits the pretrained feature extractor and overcomes the weaknesses of vanilla ER, backed up by a probabilistic analysis.
    \item Experimental results show our method outperform current state-of-the-art baselines.
\end{enumerate}

\section{Related Work}

\subsection{Continual Learning}

Mainstream continual learning strategies can be divided into three categories~\cite{review}.
\paragraph{Regularization} strategies preserve the past knowledge by directly limiting the update of the parameters so that the neural network can have stable outputs~\cite{ewc,si,mas,agscl}. The algorithms assign a penalty term for each parameter according to its importance and deviation.
\paragraph{Architecture} strategies prevent knowledge interference by assigning a task-specific module for each task and freezing shared parameters. The module can be hidden units~\cite{xdg}, parameter mask~\cite{hat} or sub-networks~\cite{den,der_yan}. In inference phase, model will switch to corresponding parameters according to task identity, or alternatively leverage the whole expanded model. 
\paragraph{Replay} (a.k.a rehearsal) strategies maintain a episodic memory to store a few past samples. During training. Experience replay (ER)~\cite{er1995,clear} takes input sampled from both of current training data and memory as if input were sampled from joint data distribution. Meta-Experience Replay (MER)~\cite{mer} combines ER with meta-learning to promote transfer and suppress interference. Gradient based sample selection (GSS)~\cite{gss} improves memory selection strategy by maximizing the diversity of gradients. Dark experience replay++ (DER++)~\cite{derpp} not only stores the ground truth in the memory but also records network's logits for replay.
\subsection{Incremental Classifier Learning}
Incremental classifier learning~\cite{icl,revisit_icl} suffers both knowledge degradation and prediction bias. iCaRL~\cite{icarl} tackles the two challenges by combining distillation loss~\cite{lwf} with experience replay. This hybrid strategy is typically used in follow-up works. However, ER can only alleviate prediction bias to some extent. Due to the memory limit, there is a severe imbalance between new classes and old classes, which encourages the classifier to output a high score for new classes. To this end, EEIL~\cite{eeil} proposes a cross-distilled loss function to retain the knowledge together with a balanced finetuning. This finetuning is replaced by a rebalanced loss in our method, which is much more efficient since it does not introduce extra training. BiC~\cite{bic} adds a linear model behind the classifier to correct the bias towards new classes. UCIR~\cite{ucir} uses cosine normalization in the last layer to avoid the imbalanced magnitude of weight vector of the classifier. They also further exploit the memory samples by a margin rank loss. SS-IL~\cite{ssil} analyzes ER from a view of gradient and proposes separated softmax to reduce inter-task interference which is contrary to our method. Among previous works, UCIR is most close to our method. UCIR also fixes the previous weights. The difference is we abolish distillation loss and only leverage a joint cross entropy loss and we equip it with a dynamic balance factor.

\section{Preliminaries}
\subsection{Problem Formulation}
Under continual learning setting, a classification problem consists of a sequence of $T$ tasks. At each time $t\in\{1,2,\cdots,T\}$, a new task arrives and model can only sample from current distribution $D_t$ while $D_i,\ 1\leq i<t$ are invisible. Due to the difficulty of class-incremental learning, this restriction is usually relieved that it is tolerable to store a small episodic memory $M$. The final goal is to learn a mapping $f:\mathcal{X}\mapsto\mathcal{Y}$ that minimizes
\begin{equation}
    \mathcal{L}=\mathbb{E}_{(x,y)\sim \mathcal{D}}\ \ell(f(x),y),\label{eq:obj}
\end{equation}
where $\mathcal{D}$ is the joint distribution of all tasks. In this paper, we assume the label set $\mathcal{Y}_t$ of each task is informed on task arriving and $\mathcal{Y}_i\bigcap\mathcal{Y}_j=\emptyset,\ \forall i\neq j$. That is to say the training is task-aware.

\subsection{A Probabilistic View of Continual Learning}\label{anal}
First, we present a analysis of what the model learns at each time $t$ from the probabilistic view. There are two typical approaches to model the data distribution with neural network. The discriminative approach models the conditional distribution $p(y|x)$ while the generative approach models the joint distribution $p(x,y)$. Most classification model belongs to the former one, since it usually enjoys better accuracy. But modeling $p(y|x)$ requires samples (original images or features or information in other forms) from all classes at the same time. By contrast, it is still feasible to train generative models even if each task contains only one class, but their classification performance usually fall behind discriminative models. In this paper, we focus on discriminative models. We assume $f(x)\triangleq g(h(x))=\text{softmax}(W^{\rm T} h(x))$, where $h$ is the frozen pretrained feature extractor, $W$ is the weight of a one-layer softmax classifier.

As tasks continually arrive, the label set is expanding and so is $W$. Note that the classifier computes probability over all seen classes. Thus we can rewrite the conditional distribution $p(y|x)$ at time $t$ as $p(y|x;\mathcal{Y}_{1:t})$. Total probability theorem gives a more detailed factorization:
\begin{equation}
    \begin{split}
        p(y|x;\mathcal{Y}_{1:t})=p(y|x;\mathcal{Y}_{1:t-1})p(y\in\mathcal{Y}_{1:t-1}|x;\mathcal{Y}_{1:t})+\\ p(y|x;\mathcal{Y}_t)p(y\in\mathcal{Y}_t|x;\mathcal{Y}_{1:t})
        \label{eq:factor}.
    \end{split}
\end{equation}
Following equation~\ref{eq:factor}, we intuitively explain the classification process as 2 phases. First, the model makes a inter-task classification to decide whether $x$ belongs to current task. Next, a intra-task classification is conducted within reduced label set. Due to $p(y\in\mathcal{Y}_{1:t-1}|x;\mathcal{Y}_{1:t})+p(y\in\mathcal{Y}_t|x;\mathcal{Y}_{1:t})=1$, the equation actually has three terms:
\begin{enumerate}
    \item $p(y|x;\mathcal{Y}_t)$. This term represents the performance on the current task. Model needs to establish this distribution at time $t$. Since model has full access to $D_t$, $p(y|x;\mathcal{Y}_t)$ can be accurately modeled.
    \item $p(y|x;\mathcal{Y}_{1:t-1})$. This term represents the performance on the previous task. Model has learned this distribution before time $t$. When model parameters update, $p(y|x;\mathcal{Y}_{1:t-1})$ may suffer knowledge degradation.
    \item $p(y\in\mathcal{Y}_{1:t-1}|x;\mathcal{Y}_{1:t})$. This term represents model's ability to distinguish samples of previous tasks from current ones. This distribution is also established at time $t$. The unavailability of previous data makes the learning of $p(y\in\mathcal{Y}_{1:t-1}|x;\mathcal{Y}_{1:t})$ challenging. 
\end{enumerate}

\begin{figure*}
    \centering
    \includegraphics[width=0.6\textwidth]{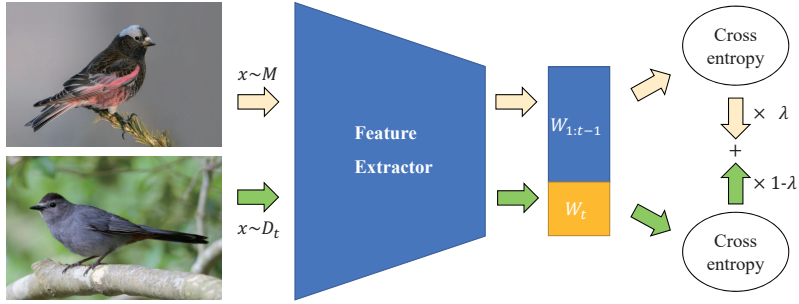}
    \caption{Overview of our model. The classifier continually expands as new task arrives. The blue parts are frozen. Only the classifier weight for the current task is allowed to update. Each step the model inputs two equal-size batches from current task and the memory respectively. The losses of two branches are integrated with a balance factor.}
    \label{fig:overview}
\end{figure*}

\subsection{Weaknesses of Vanilla ER}
Experience replay is designed to remind the model how to perform on previous tasks. The effect of ER is actually two-fold: preserving $p(y|x;\mathcal{Y}_{1:t-1})$ and establishing $p(y\in\mathcal{Y}_{1:t-1}|x;\mathcal{Y}_{1:t})$.

We next bridge the probabilistic analysis with the loss function. Vanilla ER optimizes whole weight matrix of the classifier as follows:
\begin{equation}
    \begin{split}
        &\argmin\limits_{W}\ \mathcal{L}_t^c+\mathcal{L}_t^r,\quad\text{where}\\
        &\mathcal{L}_t^c=\mathbb{E}_{x,y\sim D_t}[\ell(\sm(W^{\rm T} h(x)),y)],\\
        &\mathcal{L}_t^r=\mathbb{E}_{x,y\sim M}[\ell(\sm(W^{\rm T} h(x)),y)],
    \end{split}
\end{equation}
$\pi$ represents the softmax function and $M$ is the episodic memory. We denote $w_i$ as the $i$-th column of matrix $W$, $W_{1:t-1}$ as the columns corresponding to previous tasks and $W_t$ as the columns for current task. Assuming $\ell$ is cross entropy loss function, we can rewrite $\mathcal{L}_t^c$ as
\begin{equation}
    \mathbb{E}_{x,y\sim D_t}\left[\ell(\sm(W_t^{\rm T}h(x)),y)-\ln\frac{\sum_{i\in\mathcal{Y}_t}e^{w_i^{\rm T}h(x)}}{\sum_{i\in\mathcal{Y}_{1:t}}e^{w_i^{\rm T}h(x)}}\right],
\end{equation}
where the two terms respectively represent $p(y|x;\mathcal{Y}_t)$ and $p(y\in\mathcal{Y}_t|x;\mathcal{Y}_{1:t})$. $L_t^c$ also has a similar form:
\begin{equation}
    \mathbb{E}_{x,y\sim M}\left[\ell(\sm(W_{1:t-1}^{\rm T}h(x)),y)-\ln\frac{\sum_{i\in\mathcal{Y}_{1:t-1}}e^{w_i^{\rm T}h(x)}}{\sum_{i\in\mathcal{Y}_{1:t}}e^{w_i^{\rm T}h(x)}}\right]
\end{equation}
where the two terms respectively represent $p(y|x;\mathcal{Y}_{1:t-1})$ and $p(y\in\mathcal{Y}_{1:t-1}|x;\mathcal{Y}_{1:t})$.

In practice, although with the episodic memory, ER still suffers heavy forgetting and causes prediction bias towards classes of the current task. We attribute its poor performance to inaccurate estimates of $p(y|x;\mathcal{Y}_{1:t-1})$ and $p(y\in\mathcal{Y}_{1:t-1}|x;\mathcal{Y}_{1:t})$. Note that $M$ is very small, e.g. $|M|=200$ in our experiments, which means each class only preserves about one or two samples. Thus the classifier easily gets overfitting. Besides, in joint learning all classes are balanced sampled, but vanilla ER samples from the union set $D_t\cup M$, or samples two batches of equal size from $D_t$ and $M$ respectively at each step, which causes more exposition to the classes of the current task. This imbalance significantly hurts $p(y\in\mathcal{Y}_{1:t-1}|x;\mathcal{Y}_{1:t})$.

\begin{figure}[t]
    \centering
    \includegraphics[width=1\linewidth]{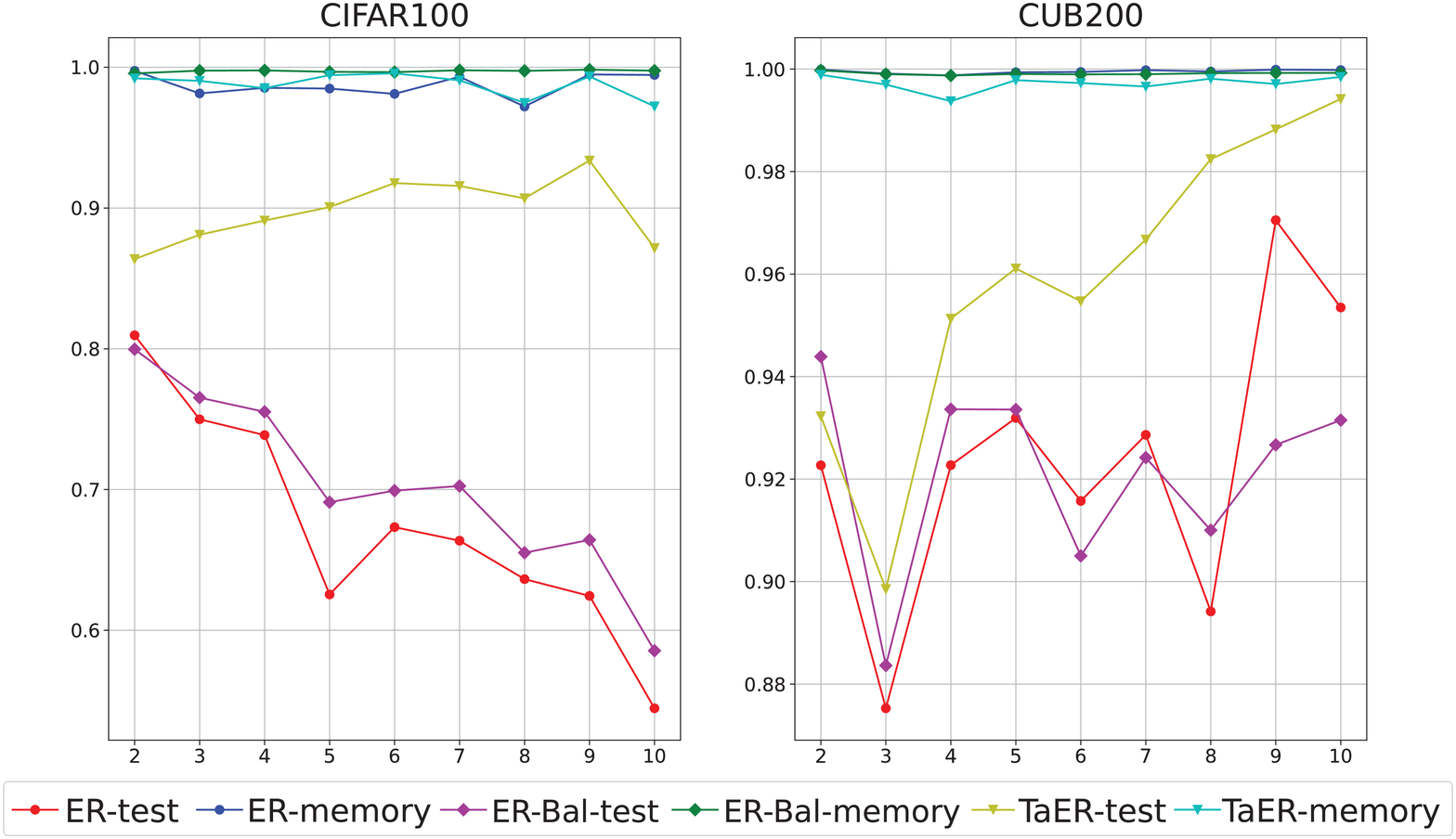}
    \caption{Estimates of $p(y\in\mathcal{Y}_{1:t-1}|x;\mathcal{Y}_{1:t})$ for each $t$ (horizontal axis). Suffixes test and memory represent the calculation of $\frac{\sum_{i\in\mathcal{Y}_{1:t-1}}\exp(w_i^{\rm T}h(x))}{\sum_{i\in\mathcal{Y}}\exp(w_i^{\rm T}h(x))}$ performed on $D_t$ and $M$ respectively. ER-Bal mean ER with balanced loss but not fixed weight. Note that classes from the current task is not included in the calculation. So the results should be close to 1 ideally. We can find all three methods correctly predicts on memory set but performance of ER and ER-Bal significantly drop on test set while TaER still works well.}
    \label{fig:prob2}
\end{figure}

\section{Task-Aware Experience Replay}
We now propose a simple but effective strategy called task-aware experience replay (TaER) that works well both for knowledge preservation and inter-task separation. It forces the model to exclusively learn the current task avoiding unnecessary parameter update while maintaining accurate boundary between old and new classes.

As shown in Figure~\ref{fig:overview}, we keep $W_{1:t-1}$ untouched since task $t$. Suppose $W_{1:t-1}$ has been well learned on $D_{1:t-1}$, there is no need to re-learn it on memory set one more time. This strategy is rarely used in the method without the aid of pretrained model, since the representation is continually learned and classifier has to adapt the changing representation. Previous works~\cite{ssil,ucir,derpp} usually leverage knowledge distillation to constrain the update of the classifier. But it is reasonable here because we assume the feature extractor has been able to extract rich feature, so the classifier no longer needs a finetuning. Frozen $W_{1:t-1}$ and $h$ brings constant $\sm(W_{1:t-1}^{\rm T}h(x))$, which means the forgetting of $p(y|x;\mathcal{Y}_{1:t-1})$ is completely eliminated. The only goal at time $t$ is to learn the classification within task $t$ and distinguish task $t$ from task $1\sim t-1$. Besides, we find that the fixed weight also reduces overfitting on estimating $p(y\in\mathcal{Y}_{1:t-1}|x;\mathcal{Y}_{1:t})$. We think this is because finetuing $W_{1:t-1}$ on $M$ will cause task boundary to collapse towards the memory sample.

To overcome class imbalance,~\cite{eeil} proposes to add balanced finetuning stage after training each task, where the model optimizes
\begin{equation}
    \mathcal{L}^b=\mathbb{E}_{(x,y)\sim M\cup\tau(D_t)}[\ell(f(x),y)]
\end{equation}
on reduced training set $\tau(D_t)$ together with the memory set. We find this procedure can be replaced by a rebalanced loss. Here we conduct replay by taking two equal-size batches from the training set and memory set respectively at each update step. Hence the model is actually optimizing
\begin{equation}\label{equal_batch}
    \mathbb{E}_{(x,y)\sim M}[\ell(f(x),y)]+\mathbb{E}_{(x,y)\sim D_t}[\ell(f(x),y)].
\end{equation}
If we further adopt a class-balanced memory management strategy, the first term in Equation~\ref{equal_batch} can be regards as an approximation of
\begin{equation}\label{eq:d1}
    \sum_{i=1}^{t-1}\left\{\frac{|\mathcal{Y}_i|}{|\mathcal{Y}_{1:t-1}|}\mathbb{E}_{(x,y)\sim D_i}[\ell(f(x),y)]\right\}.
\end{equation}
Above derivation first leverages the total probability theorem and substitute $p((x,y)\in D_i;M)$ by $\frac{|\mathcal{Y}_i|}{|\mathcal{Y}_{1:t-1}|}$. Suppose $\mathcal{D}$ is a class-balanced distribution. We factorize the equation~\ref{eq:obj} by the same way:
\begin{equation}\label{eq:d2}
    \sum_{i=1}^{t}\left\{\frac{|\mathcal{Y}_i|}{|\mathcal{Y}_{1:t}|}\mathbb{E}_{(x,y)\sim D_i}[\ell(f(x),y)]\right\}.
\end{equation}
Comparing Equation~\ref{eq:d1} and \ref{eq:d2}, we get the ratio of loss term for new and old tasks $\frac{|\mathcal{Y}_t|}{|\mathcal{Y}_{1:t-1}|}$.
It is approximately equivalent to sampling one batch from each task and take the sum. Here we give the final objective for TaER:
\begin{equation}
    \argmin\limits_{W_t}\ (1-\lambda)\mathcal{L}_t^c+\lambda\mathcal{L}_t^r,\quad \text{where } \lambda=\frac{|\mathcal{Y}_{1:t-1}|}{|\mathcal{Y}_{1:t}|}.
\end{equation}

We empirically show when the classifier is trained with TaER, the prediction bias is significantly alleviated. To estimate $p(y\in\mathcal{Y}_{1:t-1}|x;\mathcal{Y}_{1:t})$, we calculate  $\frac{\sum_{i\in\mathcal{Y}_{1:t-1}}e^{w_i^{\rm T}h(x)}}{\sum_{i\in\mathcal{Y}}e^{w_i^{\rm T}h(x)}}$ as the indicator. The experiment is conducted on 10-split CIFAR100 following the setup in section \ref{sec:exp}. As shown in Figure~\ref{fig:prob2}, both ER and TaER perfectly model the distribution on memory set but only TaER generalizes remarkably. It is worth noting that balanced loss alone is still insufficient to accurately estimate $p(y\in\mathcal{Y}_{1:t-1}|x;\mathcal{Y}_{1:t})$ unless the overfitting is overcome.

\section{Experiments}\label{sec:exp}
\begin{table*}[t]
    \centering
    \begin{tabular}{l|ccc|ccc}
    \toprule
         \multirow{2}{*}{\textbf{Method}}\quad &\multicolumn{3}{c}{ViT-S/16}&\multicolumn{3}{|c}{ResNet18}\\
         &CIFAR100 & CUB200 & CORe50 &CIFAR100 & CUB200 & CORe50\\
         \noalign{\smallskip}
         \hline\noalign{\smallskip}
         Finetune&25.00&46.57&22.40&11.50&10.78&12.58\\
         oEWC&20.41&44.30&19.77&7.85&6.31&9.69\\
         \noalign{\smallskip}
         \hline\noalign{\smallskip}
         ER&52.70&80.00&71.62&25.29&31.73&44.85\\
         iCaRL&64.06&74.87&74.57&31.49&28.94&58.36\\
         GSS&45.61&61.13&66.55&18.92&23.08&41.97\\
         DER++&56.47&81.97&75.46&22.35&30.25&51.45\\
         UCIR&52.74&\textbf{85.26}&61.75&29.20&24.54&32.27\\
         SS-IL&65.36&79.24&65.28&36.31&33.36&53.01\\
         TaER (Ours)&\textbf{71.37}& 84.47 & \textbf{78.90}&\textbf{55.50}&\textbf{55.42}&\textbf{68.85}\\
         \noalign{\smallskip}
         \hline\hline\noalign{\smallskip}
         Joint&81.50&87.00&89.43&66.47&65.86&76.44\\
        \bottomrule
    \end{tabular}
    \caption{Average accuracy after training all tasks on three benchmarks, the higher the better. The results are averaged over 3 runs.}
    \label{tab:main}
\end{table*}

\subsection{Setup}

\subsubsection{Benchmark.}
Evaluations are performed on three commonly used datasets: CIFAR100~\cite{cifar}, CUB200~\cite{cub200} and CORe50~\cite{core50}. We do not conduct experiments on ImageNet because the feature extractor used in our experiments is pretrained on ImageNet~\cite{imagenet}. There are two commonly used protocols: (i) split all classes equally into $T$ tasks~\cite{icarl}; (ii) train the model on half of the classes as the first task followed by $N$ equally split tasks and evaluate on all $T=N+1$ tasks. Under the first protocol, we conduct experiments to prove our method outperforms generic continual learning methods in linear probing a pretrained model. Besides, we use the second protocol to compare with current state-of-the-art class-incremental methods that do not use the pretrained model. $T$ is set to 10 in the former protocol. For the latter one, we set $N=5$ and $10$ and evaluate only on CIFAR100.

\subsubsection{Baselines.}
For the first protocol, we mainly compare our method against replay-based method: vanilla ER~\cite{clear}, iCaRL~\cite{icarl},  GSS~\cite{gss} and DER++~\cite{derpp}. We also compare with a regularization method oEWC~\cite{oewc} and two classifier learning methods UCIR~\cite{ucir} and SS-IL~\cite{ssil}. We do not compare with architecture-based methods here because they conflict with the linear probing setting. To better understand the performance of these continual learning methods, we present the results of naive sequential finetuning and jointly learning all tasks as reference. They respectively represent the lower bound and upper bound of this learning scenario. For the second protocol, we compare with architecture-based methods DER~\cite{der_yan}, AANets~\cite{aanet} and FAS~\cite{fas}. The former two methods are also replay-based.

\subsubsection{Implementation Details.}
For the methods used in protocol (i), we choose ViT-S/16~\cite{vit} and ResNet18 as the frozen pretrained extractor (pretrained on ImageNet) and produce the results using their public codes. We train the classifier with a SGD optimizer with batch size 32. Since there is no need to learn the representation, one epoch is sufficient for CIFAR100 and CORe50. As CUB200 contains relatively few samples, it needs more training, so we set epoch as $10$ for it. For our method, ER and finetune, learning rate is set to $0.1$ without momentum. For all experiments, no data augmentation is applied. The size of the episodic memory is fixed as 200. We use greedy balancing sampling strategy~\cite{gdumb} to manage the memory except GSS and UCIR since the management strategy is part of their methods. For the baselines used in protocol (ii), we directly use the results reported in their papers. Following their settings, the size of the episodic memory is set to 2000. As the first task contains 50 classes, we initialize the classifier weight with the mean of the feature of the corresponding class computed on training set to accelerate learning.

\subsubsection{Metrics.} 
Following the previous work~\cite{rwalk}, we report average accuracy $A_T=\frac{1}{T}\sum_{k=1}^Ta_{T,k}$, where $a_{T,k}$ represents the accuracy evaluated on task $k$ after training $T$ tasks. Under protocol (ii), due to the imbalanced number of classes in each task, the commonly used metric is average incremental accuracy~\cite{ucir} $AIA_T=\frac{1}{T}\sum_{k=1}^T\alpha_k$ where $\alpha_k$ is the accuracy evaluated on all seen classes at training phase $k$.

\begin{table}[t]
    \centering
    \begin{tabular}{l|ccccc}
    \toprule
        \multirow{2}{*}{N}\quad & DER & AANET & FAS & TaER&TaER \\
         &RN18&RN32&RN32&RN18&ViT-16/S\\
         \noalign{\smallskip}
         \hline\noalign{\smallskip}
         5 & 73.21 & 67.59&65.44&68.86&\textbf{80.68}\\
         10 & 72.81 & 65.66&62.48&68.27&\textbf{80.15}\\
        \bottomrule
    \end{tabular}
    \caption{Average incremental accuracy on CIFAR100 where the first class contains half of classes. "RN" is the abbreviation of ResNet.}
    \label{tab:cifar100-B50}
\end{table}
\subsection{Evaluation for Linear Probing Setting}


As shown in Table~\ref{tab:main}, TaER outperforms all other baselines using the ResNet18 backbone. For the ViT-S/16 backbone, TaER achieves better average accuracy than all other baselines on CIFAR100 and CORe50 and is second best on CUB200. Although UCIR get a slightly higher accuracy on CUB200, its performance on the other two benchmarks are much lower than ours. It is notable that the best result among baselines on three datasets comes from three different methods. No baseline consistently outperforms the others. Overall, TaER achieve the best performance and has better generalization.

As for other methods, non-replay methods SGD and oEWC performs worst because they cannot distinguish across different tasks. This is especially notable for ResNet18 features where the model can only work on the last task. By comparison, replay methods achieve unbridgeable improvement even each class is assigned about two memory samples. We observe ER is competitive among these baselines which consistently outperforms GSS. It seems that the gradient-based sample selection strategy does not work in this setting. DER++, UCIR and SS-IL respectively beat the other baselines on three datasets. The generic replay method DER++ is the only baseline consistently outperforms ER. The characteristics of the dataset seem to have significant influence on UCIR and SS-IL.

\subsection{Comparison with Other SOTA Baselines}
As shown in Table~\ref{tab:cifar100-B50}, ViT-based TaER outperforms all other baselines again. FAS performs the worst because it does not store samples or features. DER outperforms the other two baselines drastically because it uses ResNet18 rather than reduced ResNet32. When switched to ResNet18 backbone, TaER only outperforms AANET and FAS and falls behind DER, because DER can adapt features on downstream tasks which benefits more from large episodic memory. Although the ViT-S/16 used in TaER is larger than ResNet18 and reduced ResNet32, its trainable parameters are much fewer. Actually their training costs are very close. So, these results are comparable. Moreover, the architecture-based methods has two drawbacks. First, architecture-based method is usually designed for a specific network architecture (e.g ResNet) which makes it hard to apply on other models (e.g. ViT). Second, architecture-based method causes model size to increase as task continually arrives. The amount of increment depends on the size of the base model. The larger the model, the larger the size increment. These two drawbacks impede these SOTA baselines to cooperate with pretrained models. But our TaER is compatible with pretrained models of any architecture and any size. The memory increment only depends on the number of classes and dimension of features, which are very small relatively to the model size for almost all applications. We argue the applicability of pretrained model is the advantage of our method and should be take into account when making comparison.

\subsection{More Analysis and Discussion}
Considering that all methods perform much better using ViT-S/16, we only discuss the performance using ViT-S/16 as the feature extractor in the following.
\begin{figure}[t]
    \centering
    \includegraphics[width=\linewidth]{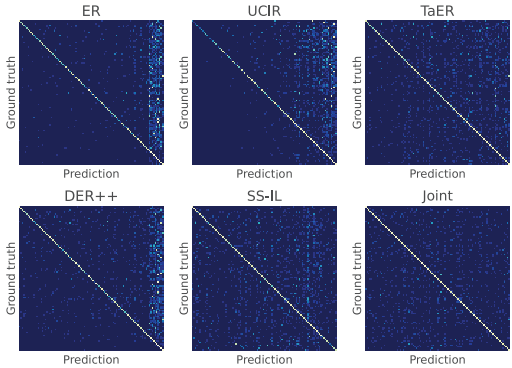}
    \caption{The visualization of the confusion matrix after 10 tasks trained on CIFAR100. The pixel in $p$-th row and $q$-th column represents the ratio that model classifies sample with ground truth $p$ into label $q$. Higher luminance means larger ratio.}
    \label{fig:bias}
\end{figure}
\subsubsection{Prediction Bias.}
To verify that TaER alleviates prediction bias, we compare the confusion matrix of our method with other baselines. As shown in Figure \ref{fig:bias}, we can find ER, UCIR and DER++ have strong tendency to classify the old classes into the most recent classes. This is because of aforementioned class-imbalanced sampling and overfitting on memory set. By comparison, TaER shows much less prediction bias and its prediction mainly concentrates on the diagonal. SS-IL shows even less bias towards the last task, but we can find the diagonal at right bottom corner is much darker. It has a reversed bias towards the previous tasks when predicting on the most recent task, which we also do not expect. TaER pursues overall inter-task separation and thus have a better performance.

\begin{figure}[t]
    \centering
    \includegraphics[width=\linewidth]{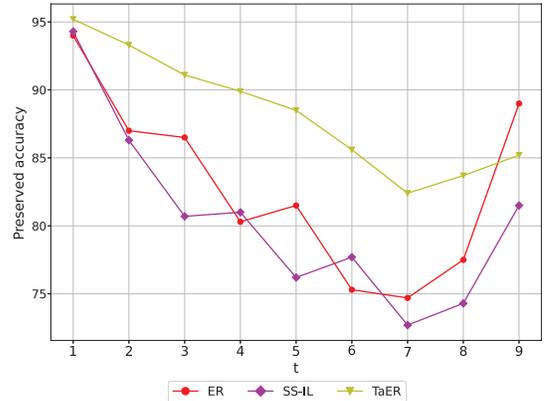}
    \caption{Preserved accuracy evaluated after training 10 tasks on CIFAR100.}
    \label{fig:kp}
\end{figure}

\subsubsection{Knowledge Preservation}
In section~\ref{anal} we have decoupled catastrophic forgetting into two parts. Besides the prediction bias that we have discussed, knowledge degradation is important as well. To measure the knowledge preservation, we define preserved accuracy $P_t^T$ which means the accuracy computed on $D_t$ over the label set $\mathcal{Y}_{1:t}$ after training all $T$ tasks. This definition is different from $a_{T,k}$ where the classification is conducted over the full label set $\mathcal{Y}_{1:T}$. By this way, we eliminated the performance drop caused by increasing classes. The results are shown in Figure~\ref{fig:kp}. The curve has a tendency to decrease at the early stage but turn to increase at last. The high preserved accuracy on the starting tasks is easy to understand since there are only a few classes to predicted. As for the most recent tasks, they have been just trained and have not experienced much forgetting. We can find TaER does better in knowledge preservation. This is credited to the weight fixing strategy. It keeps all the knowledge learned at each training phase.
\begin{small}
    \begin{table}[t]
    \centering
    \begin{tabular}{lccc}
    \toprule
         \textbf{Objective} & CIFAR100 & CUB200 & CORe50 \\
         \noalign{\smallskip}
         \hline\noalign{\smallskip}
         $\argmin\limits_{W}\mathcal{L}_t^c\!+\!\mathcal{L}_t^r$&52.70&80.00&71.62\\
         $\argmin\limits_{W_t}\mathcal{L}_t^c\!+\!\mathcal{L}_t^r$&57.47&79.87&70.49\\
         $\argmin\limits_{W}\bar\lambda\mathcal{L}_t^c\!+\!\lambda\mathcal{L}_t^r$&59.73&80.50&76.33\\
         $\argmin\limits_{W_t}\bar\lambda\mathcal{L}_t^c\!+\!\lambda\mathcal{L}_t^r$&\textbf{71.37} & \textbf{84.47} & \textbf{78.90}\\
        \bottomrule
    \end{tabular}
    \caption{Ablation study for TaER. $\bar\lambda=1-\lambda$. We report average accuracy on three datasets.}
    \label{tab:ablation}
\end{table}
\end{small}

\subsubsection{Ablation Study.}
We provide an ablation study on the three benchmarks. As shown in Table \ref{tab:ablation}, both the weight fixing strategy and loss rebalance factor boost the average accuracy for CIFAR100. Fixing weight alone (the second row) seems not to bring improvement on CUB200 and CORe50. We think this is because knowledge degradation and overfitting on memory set is not so serious as CIFAR100. The results of ER (the first row) support this view, especially on CUB200 where the performance of ER has been very close to the upper bound. On CORe50, loss rebalancing factor (the third row) contributes the most improvement. We speculate this is because prediction bias is primary challenge on CORe50.  From an overall view, loss rebalancing factor tends to play a more important role than the weight fixing strategy. When combing the two components, TaER consistently beat other variations.

\begin{figure}[t]
    \centering
    \includegraphics[width=\linewidth]{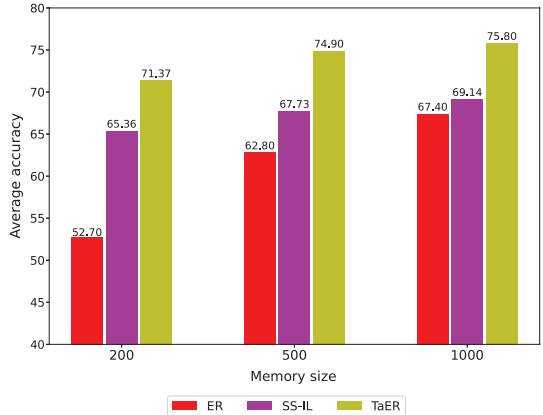}
    \caption{Average accuracy evaluated with different memory size.}
    \label{fig:memory_acc}
\end{figure}

\subsubsection{The Effect of Memory Size.}
We provide the results about the effect of memory size in Figure~\ref{fig:memory_acc}. It is apparent that larger memory benefits all methods, especially ER. As we have analyzed before, small memory will cause heavy overfitting. The size of the memory is critical to the performance of ER. Our method also benefits a lot from the larger memory for the same reason because weight fixing strategy can only alleviate overfitting but not get rid of it. The estimate of $p(y\in\mathcal{Y}_{1:t-1}|x;\mathcal{Y}_{1:t})$ still relies on the previous samples. As for SS-IL, the incremental is not so significant. It is noteworthy that SS-IL only uses the memory to re-learn the past tasks rather than jointly classify with the current task. We think this is because the memory values more for inter-task separation than knowledge preservation.

\begin{table}[t]
    \centering
    \begin{tabular}{lccc}
    \toprule
         N\quad & ER & SS-IL & TaER \\
         \noalign{\smallskip}
         \hline\noalign{\smallskip}
         5 & 59.40&69.96&\textbf{72.66}\\
         10 & 52.70&65.36&\textbf{71.37}\\
         20 & 51.63&53.43&\textbf{71.09}\\
         50 & 48.16&47.37&\textbf{66.62}\\
        \bottomrule
    \end{tabular}
    \caption{Average incremental accuracy on CIFAR100 of different task numbers.}
    \label{tab:cifar100-T}
\end{table}

\begin{figure}
    \centering
    \includegraphics[width=\linewidth]{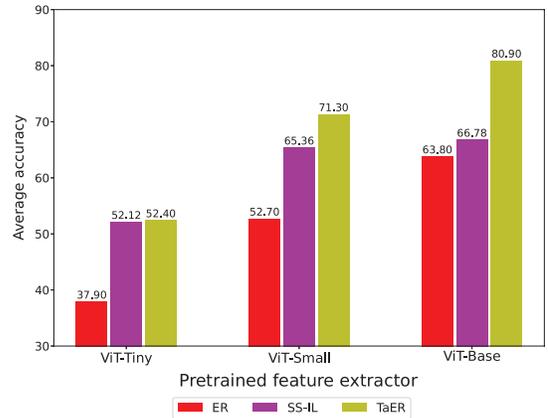}
    \caption{Average accuracy evaluated with different pretrained feature extractor.}
    \label{fig:model_acc}
\end{figure}

\subsection{The Effect of Task Numbers}
To understand how our method performs on longer or shorter learning sequence, we additionally provide results on 5-split, 20-split and 50-split CIFAR100. Longer sequence makes knowledge preservation and inter-task separation even harder. But our model still maintains a remarkable average accuracy as shown in Table~\ref{tab:cifar100-T}. The performance for 5, 10 and 20 tasks are quite close. The only apparent drop happens in 50-split setting, where the model can only see two classes at each task. By comparison,  both ER and SS-IL experience a drastic performance dropping as task number increases.

\subsection{The Effect of Feature Extractor}
It is worth noting that TaER is designed to train a classifier built on pretrained model. The capacity of the pretrained model is critical to TaER's performance. To further understand the effect of feature extractor, we conduct experiments with ViT-Tiny, ViT-Small and ViT-Base (the amount of parameters are 5.5M, 21.7M and 85.8M respectively) on CIFAR100. As shown in Figure~\ref{fig:model_acc}, larger model benefits all methods. Our method achieves the highest accuracy on the largest model by a margin more than 14\% over the others, which proves that TaER is good at exploiting large pretrained model. The improvement of accuracy is strongly correlated with the increment of model complexity. It is conceivable that more powerful pretrained model will bring even larger improvement.

\section{Conclusions}

In this paper, we step towards large pretrained model based continual learning and focus on linear probing setting. We analyze the weaknesses of vanilla ER, a most widely used strategy, from a probabilistic view and accordingly decouple the catastrophic forgetting into knowledge degradation and prediction bias two parts. To overcome ER's weaknesses we propose a novel ER strategy called task-aware experience replay and empirically verify that TaER works as we expect. Extensive experiments prove TaER outperforms current SOTA baselines on several benchmarks. In addition, we reveal the potential of pretrained model to solve continual learning problems. With the aid of large pretrained model, our TaER has shown dominant superiority to the classic SOTA methods even just by linear probing. 

\bibliography{mybib}

\end{document}